\documentclass[journal]{IEEEtran}

\ifCLASSINFOpdf
  \usepackage[pdftex]{graphicx}
\else
\fi

\usepackage{amsmath}
\usepackage{amssymb}
\usepackage{bm}

\usepackage{subcaption}
\usepackage{colortbl}

\usepackage{tikz}
\usepackage{textcomp}
\usepackage{hyperref}

\newcommand{\namee}{CA-EEGNet}
\newcommand{\namew}{CA-EEGWaveNet}

\newcommand{\norm}[1]{\left\lVert#1\right\rVert}

\begin{document}
\title{A Composable Channel-Adaptive Architecture for Seizure Classification}

\author{Francesco~S.~Carzaniga, 
  Michael~Hersche,
  Kaspar~A.~Schindler, and~Abbas~Rahimi%
\thanks{
F. Carzaniga is with IBM Research -- Zurich and with the Department of Neurology, Inselspital, Sleep-Wake-Epilepsy-Center, Bern University Hospital, Bern University, Bern, Switzerland.}%
\thanks{M. Hersche and A. Rahimi are with IBM Research -- Zurich.}%
\thanks{K. Schindler is with the Department of Neurology, Inselspital, Sleep-Wake-Epilepsy-Center, Bern University Hospital, Bern University, Bern, Switzerland.}}

\IEEEpubid{DOI: 10.1109/JBHI.2025.3587103~\copyright~2025 IEEE}

\maketitle

\begin{abstract}

\textit{Objective:} We develop a channel-adaptive (CA) architecture that seamlessly processes multi-variate time-series with an arbitrary number of channels, and in particular intracranial electroencephalography (iEEG) recordings.
\textit{Methods:} Our CA architecture first processes the iEEG signal using state-of-the-art models applied to each single channel independently. The resulting features are then fused using a vector-symbolic algorithm which reconstructs the spatial relationship using a trainable scalar per channel. Finally, the fused features are accumulated in a long-term memory of up to 2 minutes to perform the classification. Each CA-model can then be pre-trained on a large corpus of iEEG recordings from multiple heterogeneous subjects. The pre-trained model is personalized to each subject via a quick fine-tuning routine, which uses equal or lower amounts of data compared to existing state-of-the-art models, but requiring only 1/5 of the time.
\textit{Results:} We evaluate our CA-models on a seizure detection task both on a short-term ($\sim$20 hours) and a long-term ($\sim$2500 hours) dataset. In particular, our CA-EEGWaveNet is trained on a single seizure of the tested subject, while the baseline EEGWaveNet is trained on all but one. Even in this challenging scenario, our CA-EEGWaveNet surpasses the baseline in median F1-score (0.78 vs 0.76). Similarly, CA-EEGNet based on EEGNet, also surpasses its baseline in median F1-score (0.79 vs 0.74).
\textit{Conclusion and significance:} Our CA-model addresses two issues: first, it is channel-adaptive and can therefore be trained across heterogeneous subjects without loss of performance; second, it increases the effective temporal context size to a clinically-relevant length. Therefore, our model is a drop-in replacement for existing models, bringing better characteristics and performance across the board.

\end{abstract}

\begin{IEEEkeywords}

Deep learning, vector-symbolic architectures, time-series, epilepsy, electroencephalography, convolutional neural networks

\end{IEEEkeywords}

\IEEEpeerreviewmaketitle

\section{Introduction}
\IEEEPARstart{M}{ulti-variate} time-series are an ubiquitous source of data, generated and processed everyday in many research and industry environments. Deep learning (DL) represents one of the most successful methods for analyzing such large quantities of data, and has been widely adopted.
In the medical domain, DL models have been used to solve a variety of tasks involving electroencephalograms recorded non-invasively with extra-cranial electrodes (EEG) signals~\cite{Liu2017, Song2020, Zhao2021, Rajpurkar2022, Wei2022, Heinrichs2023, Ho2023}, and invasively with intracranial electrodes (iEEG). 
EEG signals are often recorded in hospitals for diagnostic purposes, but their use extends into research and commercial environments~\cite{Casson2019,Arico2020} as well. 
As an example, one of the most compelling uses of iEEG in the clinic is the detection of seizures and the precise localization of seizure onset regions in subjects suffering from pharmacoresistant epilepsy~\cite{Kuhlmann2018, Craik2019}.
However, EEG data remains highly challenging for DL models. In particular, they need to be trained from scratch for each subject and are not capable of generalizing to unseen clinical recording setups, such as different numbers of channels and electrode locations. Large Transformer-based models~\cite{Yuan2024} have partially addressed this limitation, but their size and the resources necessary to train and deploy them make them impractical for clinical use.
Moreover, the amount of data that these models consider when evaluating the occurrence of a seizure, i.e., the context window, is severely restricted when compared to a human expert's. This limits their overall performance.
In practice, these limitations make such models burdensome to apply for general real-world use and have hindered their widespread adoption.
We highlight three main issues affecting DL models for multi-variate time-series in general, emphasizing their role in EEG.

\textbf{Fixed channel setup.} First, DL models often need a fixed number of channels to properly process the time-series recordings. Moreover, the setup of the channels also needs to remain fixed, as the models cannot easily generalize to unseen configurations.
Neither of these two requirements can be satisfied across heterogeneous subjects, whose recording electrodes are personalized and thus often vary in number and location.
This issue is especially relevant for iEEG recordings, since their setup is guided by clinical needs only. Nonetheless, iEEG remains the preferred data source for training DL models due to its higher signal-to-noise ratio.
Thus, DL models are often not capable of generalizing across heterogeneous subjects~\cite{Burrello2019,Wang2023} and need to be trained from scratch for each subject. This significantly increases the time and computational resources needed to deploy a fully functioning model.

\IEEEpubidadjcol

\textbf{Limited data sources.} Second, by training only on the data from a single subject, current DL models are precluded from using the large amounts of EEG signals that have been already collected~\cite{Burrello2018, Burrello2019,Shoeb2010}.
Typically, EEG signals are collected by placing electrodes on the scalp of the subject. Therefore, such EEG recordings are not invasive, and they are commonplace in many institutions, so EEG data is abundant. The availability of EEG data is also expected to increase significantly, given the recent developments of wearable EEG systems~\cite{Yilmaz2024} for ultra long-term recordings. However, non-invasively recorded EEG suffers from limited signal-to-noise ratio and low spatial and temporal resolutions. 
iEEG surpasses these limitations by implanting electrodes directly onto or even into the subject's brain. As a consequence, iEEG is markedly more invasive and its availability is limited. 
Therefore, the overall performance of DL models on iEEG suffers as a consequence of this limited availability, and would greatly benefit from being able to pre-train on larger heterogeneous datasets which are more widely available.

\textbf{Short context window.} Third, when detecting seizures in an iEEG signal, current DL models use context windows on the order of a few seconds to reduce their computational complexity. In contrast, human experts often analyse minutes of data in order to make a decision.
This limits the scale at which DL models can operate, and makes them unable to properly assess more unusual seizure presentations, which can last for multiple seconds to minutes or be closely matched to the near past signal.
Once again, the seizure detection performance suffers as a consequence.

To address the limitations outlined above and bring DL models closer to practical application in the clinic, we introduce a channel-adaptive (CA) architecture. Our CA architecture includes a Fusion-enabled pre-training to allow training on any dataset, and a temporal memory to increase the effective context size. In particular, it combines:
\begin{itemize}
    \item Fusion of an arbitrary number of channels into a single representation, which surpasses the \textbf{fixed-channel setup} limitation;
    \item The Fusion also allows for training with larger datasets composed of multiple heterogeneous subjects with different spatial setups and number of channels, which surpasses the \textbf{limited data sources} limitation;
    \item Union of multiple short signal windows to generate an integrated classification over minutes of signal, which surpasses the \textbf{short context window} limitation;
    \item Seamless integration of existing state-of-the-art models, with no limitations to compatibility with future models.
\end{itemize}
Our CA architecture integrates all the channels of an arbitrary iEEG recording while implicitly preserving the spatial structure of the signal.
Our novel Fusion schema is enabled by holographic reduced representations (HRR)~\cite{Plate1995}, and allows our model to be used across different subjects and clinical setups without any modifications to the architecture.
Using HRR, we develop a learnable and maximally efficient fusion of the channels, with only a single parameter per channel to encode the channel setup.
Moreover, our architecture uses two levels of processing---one short-term and an aggregate long-term---to greatly increase the context window with minimal impact to the model size and complexity, which makes seizure detection possible at a clinically relevant scale of up to minutes of signal.
We make use of the characteristics of our CA architecture by:
\begin{itemize}
    \item pre-training on a large multi-subject corpus of iEEG recordings with 32--128 electrodes;
    \item fine-tuning on a few seizures for a specific subject;
    \item deploying the personalized model with 5$\times$ fewer epochs required by a full retraining, and achieving state-of-the-art performance.
\end{itemize}
This training schema increases the amount of data at our disposal by 5 to 15 times, depending on the pre-training setup. 
Finally, our model achieves state-of-the-art results in the seizure detection task even in a challenging realistic scenario, where our model only has access to a single subject-specific seizure. This is in contrast to existing models that required multiple subject-specific seizures to train effectively.

\section{Related work}

\textbf{EEG classification.} DL models have been developed to solve a variety of tasks using iEEG and EEG, such as motor imagery~\cite{Schirrmeister2017,Lawhern2018}, sleep scoring~\cite{Vilamala2017,Palotti2019,Perslev2021}, seizure detection~\cite{Kuhlmann2018,Cho2020,Thuwajit2022}, and others~\cite{Rajpurkar2022,Du2022,Lee2022}. In particular, EEGNet~\cite{Lawhern2018} and EEGWaveNet~\cite{Thuwajit2022} have emerged as state-of-the-art solutions in many instances. For this reason, we adopt them as our base models. However, their context windows are limited to 1--5 seconds. Moreover, no DL model has shown the capability of cross-subject training with subjects that have a heterogeneous channel setup, as is typically the case for iEEG datasets. In contrast, our model is channel-adaptive, thus able to train on multiple subjects in parallel, and achieves an effective context window of up to 105 seconds with greater overall performance.

\noindent\textbf{Channel fusion in bio-signal classification.} Channel fusion has been used for the classification of bio-signals~\cite{Burrello2018,Moin2021} using fixed bipolar key vectors and variable value vectors. The use of fixed bipolar vectors, however, is non-differentiable and limits their flexibility when the number and the location of the electrodes vary considerably. In contrast, we use a fully differentiable, learnable schema with only one parameter per channel to represent the relevant information content of the spatial dimension.

\begin{figure*}[t]
    \centering
    \includegraphics[width=.9\textwidth]{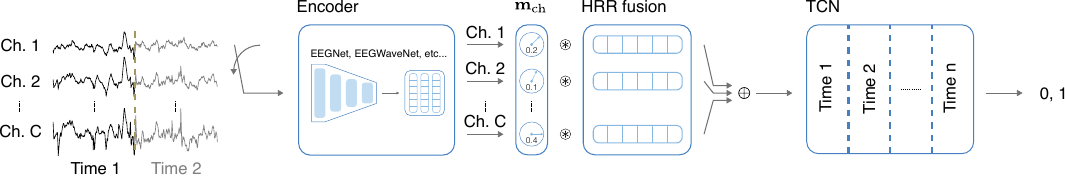}
    \caption{Architecture of the CA classifier. The classifier consists of three components. The Encoder component acts channel-wise and extracts short-term temporal features from each channel. The holographic (HRR) Fusion component combines the channel-wise features while preserving the spatial structure. Finally, the Memory component (TCN) extracts the relevant long-term temporal features and outputs the classification decision.}
    \label{fig:nvsa_arch}
\end{figure*}

\section{Channel-adaptive (CA) architecture}

This section introduces the main contribution of this work, a CA architecture with interchangeable base components (see Fig.~\ref{fig:nvsa_arch}) that combines an arbitrary number of channels while preserving the relevant spatial information content, thus enabling channel-adaptive training. We pinpoint three distinct and intrinsic characteristics of an iEEG signal that can help identify a seizure, and associate to each a component of our architecture (see Table~\ref{tab:ccc_arch}). 

\begin{table}[hbt]
    \centering
    \begin{tabular}{l l l}
        iEEG Feature    & Component                & Model \\\hline
        Short-term & Encoder         & EEGWaveNet            \\
        Long-term  & Memory  & TCN               \\
        Spatial    & Fusion & HRR              
    \end{tabular}
    \caption{Our composable CA architecture contains three components, each attending to a different aspect of the signal. We identify three features of interest: short-term, long-term, and spatial. Each feature is assigned a best-performing model.}
    \label{tab:ccc_arch}
\end{table}

An EEG signal $X \in \mathbb{R}^{C\times T}$ has $C$ channels and a duration $T = t\times f_s$ of $t$ seconds at a sampling frequency of $f_s$. The signal is first divided into short windows (on the order of seconds) and then fed to the Encoder component channel-by-channel to extract short-term features. The channel-by-channel feature vectors are fused together through our novel HRR-based learnable schema to recover the original spatial structure of the signal while producing a single representation for further processing. Finally, multiple short-term fused representations are aggregated into a long-term context window, which the Memory component uses to produce the final classification.

We choose a composable approach to allow the reuse of the architecture following the development of newer Encoder or Memory models, to preserve the channel-adaptive training schema and the fusion interface. To identify our CA model from its components, we prefix CA to the name of the Encoder model. In particular, we will consider \namee\ and \namew.

\textbf {Encoder.} First, we divide the signal ($X$) channel-wise into short patches $\bm{x}_{i, j} \in \mathbb{R}^W, \; i \in {1, \dots, C}, \; j \in {1, \dots, \frac{T}{W}}$, with the patch size $W$ on the order of a few seconds. The patches serve as the input to the Encoder model, that has the objective of extracting temporal features on the short time scale. In particular, the Encoder model is replaceable as improved solutions emerge, as long as it provides some output features $\bm{p}_{i, j} \in \mathbb{R}^d$.  This allows us to keep the same architecture but include the benefits of different models at the same time. In fact, we showcase this composability by testing the CA classifier using EEGNet~\cite{Lawhern2018} and EEGWaveNet~\cite{Thuwajit2022} as Encoder models. To handle an arbitrary number of channels, the encoder processes each $\bm{x}_{i, j}$ independently into an output vector $\bm{p}_{i, j}$, ignoring any inter-channel interaction. 
Thus, we can adapt to the specific iEEG setup and will process the spatial information content of the signal in the downstream Fusion component.

\textbf{Fusion.} The objective of the Fusion component is to integrate the spatial information content of the signal. Given input feature vectors $\bm{p}_{i, j}$, we aim to construct for each time point a single integrated representation  $\bm{f}_{j} \in \mathbb{R}^d, \; j\in \{1,\dots, T_W\}$ which contains the information we need. To preserve the spatial interactions between the channels, we develop a novel, fully trainable fusion schema based on holographic reduced representations (HRR~\cite{Plate1995}), a family of vector symbolic architectures~\cite{Gayler2003, Kanerva2009}. We present here the main steps of the Fusion component, for more detailed information see App.~B. First, we choose a random unitary (in $\mathbb{C}$) basis vector $\bm{v} \in \mathbb{R}^d$. Then, we track the relationship between the channels through a learnable vector $\bm{m}_{\text{ch}}\in \mathbb{R}^C$, with one value per channel. $\bm{m}_{\text{ch}}$ is a maximally efficient representation of the spatial information contained in the iEEG signal. The coefficients of $\bm{m}_{\text{ch}}$ indicate the angle of rotation of the basis vector in Fourier space, i.e.
\begin{equation}
    \text{rot}(\bm{v}, r) = \mathcal{F}^{-1}(\mathcal{F}(\bm{v})^{r}),
\end{equation}
where we take the element-wise power of each component of $\mathcal{F}(\bm{v})$ by an angle $r$ from $0$ to $1$ (which indicates a rotation by $2\pi$ radians). In particular, the closer the $\bm{m}_{\text{ch}}$ values of two channels are, the stronger the relationship between them.  To obtain the final integrated vector across all channels, we apply the HRR schema:
\begin{equation}
\bm{f}_j = \sum_{i}^C \bm{p}_{i, j} \circledast \text{rot}(\bm{v}, \bm{m}_{\text{ch}}^i),
\end{equation}
where $\circledast$ is the circular convolution acting as the binding operator. The HRR schema ensures that similar channels constructively interfere in the latent space while different channels do not interfere, thanks to the binding that maps them to separate protected subspaces. Moreover, the relationship between the channels is preserved as the angular distance between the components of $\bm{f}_j$, which can then be processed by the downstream components. Therefore, through the Fusion component, we can process any number of channels without loss of generality or representation power.

\textbf{Memory.} The Encoder and Fusion components extract the required spatial short-term temporal information. However, neurologists take into consideration long-term temporal dependencies on the order of minutes when making their decisions regarding seizures. Therefore, we believe that an effective seizure detection architecture must obtain long-term temporal information as well. To do this, we equip our CA classifier with a stack that collects multiple integrated patches $\bm{f}_{j}$ to form a long-term memory $\{\bm{f}_j, \dots, \bm{f}_{j+M}\}$ with length $M = 14$. Effectively we increase the context available to our network from $W$ to $W\times M$. We use a patch length of 7.5\,s to yield an effective context length of 105\,s. We process the memory stack using a temporal convolutional network (TCN)~\cite{Lea2016}, which avoids the computational pitfalls of traditional recurrent neural networks. The TCN component maintains high efficiency while processing long sequences of iEEG signal.

\section{Training setup}\label{sec:training}

We train all our CA models following a two-stage approach. First we pre-train, then we fine-tune. To assess the performance of our CA classifier we train and test on all the subjects with a rigorous procedure, by fine-tuning each model in a leave-one-out cross-validation manner.

\textbf{Pre-training.} We pre-train our CA model on a large amount of heterogeneous data to increase its ability to generalize across subjects. For each testing subject in both the Short-term~\cite{Burrello2018} and Long-term~\cite{Burrello2019} SWEC iEEG datasets, we pre-train a separate model on a number of subjects. Crucially, the subject the model is trained for is excluded from the pre-training selection, and is only used during fine-tuning and testing. In particular, we pre-train on 5, 10, and 15 subjects. As we are able to collect data from multiple subjects, the model can be trained with optimal data efficiency and does not suffer from a lack of data availability. Moreover, the selection of subjects is random to avoid bias. The model is pre-trained until the training F1-score plateaus, for a minimum of 25 and a maximum of 50 epochs. To guarantee uniform training coverage, we select data from a random pre-training subject at every batch. Depending on the number of pre-training subjects and the hardware, this stage can last from a few hours to a few days. On a Nvidia A100 80GB GPU we achieve a maximum pre-training time of $<$6 hours, with 15 subjects pre-training.

\textbf{Fine-tuning.} During the fine-tuning stage, only the testing subject is considered. Some of their seizures are fed to the model, mainly to train the spatial Fusion component, which is subject-dependent. There are two fine-tuning strategies:
\begin{itemize}
    \item Leave-one-out cross-validation (LOOC), where each model is fine-tuned on all but one seizures of the testing subject, resulting in a number of sub-models equal to the number of seizures of the subject
    \item Leave-all-but-one-out cross-validation (LABOC), where each model is fine-tuned on only one seizure of the testing subject, resulting again in a number of sub-models equal to the number of seizures of the subject.
\end{itemize}
This guarantees the validity of our results in a variety of conditions, from having equivalent data to the baselines to having the least amount of data possible. Each model is fine-tuned for a maximum of 10 epochs.

As a baseline, we also train EEGNet and EEGWaveNet models following their respective original training schema, using the LOOC regime without pre-training. In fact, EEGNet and EEGWaveNet cannot be trained using the LABOC schema, as the amount of data is insufficient and the performance is poor. In contrast, our CA models perform at the state-of-the-art even with such a limited amount of subject-specific data.

\subsection{Parameters}

To obtain the best performing models, both for our CA models and the two baselines, we perform a round of hyper-parameter optimization with each architecture using the Async Successive Halving Algorithm (ASHA) scheduler. For both \namee\ and \namew, 5 random subjects are selected for pre-training and 2 more random subjects are selected for validation. For the baseline models, all but one of the seizures of one random subject are selected for training, and the other is left for validation. In every instance we test at least 100 hyper-parameter combinations and choose the one with highest validation F1-score. The optimisation is performed on the Short-term dataset, and carried over to the Long-term dataset.

Remarkably, our CA architecture offers the best performance with the same set of hyper-parameters for both Encoder choices. The resulting sizes for the baselines models are 2M parameters and 227K parameters for EEGNet and EEGWaveNet respectively. For our CA models, the sizes are 1.7M and 995K for \namee\ and \namew\ respectively.

Table~\ref{tab:hyper_cac} reports the hyper-parameter selection for \namee\ and \namew, while Table~\ref{tab:hyper_base} for the baselines. 

\begin{table}
    \centering
    \begin{tabular}{lll}
    & \textbf{\namee}        & \textbf{\namew}                                    \\ 
    \hline
    Learning rate                                         & $5.5 \times 10^{-4}$ & $5.5 \times 10^{-4}$  \\
    Window                                                & 7.5\,s            & 7.5\,s                                            \\
    Stride                                                & 1\,s            & 1\,s                                            \\
    Memory                                                & 14            & 14                                             \\
    Pointwise filters                                     & 2            & N.A.                                          \\
    Spatial filters                                       & 16            & N.A.                                          \\
    Spectral filters                                      & 8            & N.A                                          
    \end{tabular}
    \caption{Hyper-parameter final selection for \namee\ and \namew\ models.}
    \label{tab:hyper_cac}
\end{table}

\begin{table}
    \centering
    \begin{tabular}{lll}
                                                          & \textbf{EEGNet}        & \textbf{EEGWaveNet}                                    \\ 
    \hline
    Learning rate                                         & $1.0\times 10^{-2}$ & $3.0 \times 10^{-4}$  \\
    Window                                                & 5\,s            & 5\,s                                            \\
    Stride                                                & 1\,s            & 1\,s                                            \\
    Pointwise filters                                     & 20            & N.A.                                          \\
    Spatial filters                                       & 16            & N.A.                                          \\
    Spectral filters                                      & 26            & N.A                                          
    \end{tabular}
    \caption{Hyper-parameter final selection for baseline models.}
    \label{tab:hyper_base}
\end{table}

\begin{figure*}
\begin{subfigure}{.45\linewidth}
    \centering    
    \includegraphics[width=\linewidth]{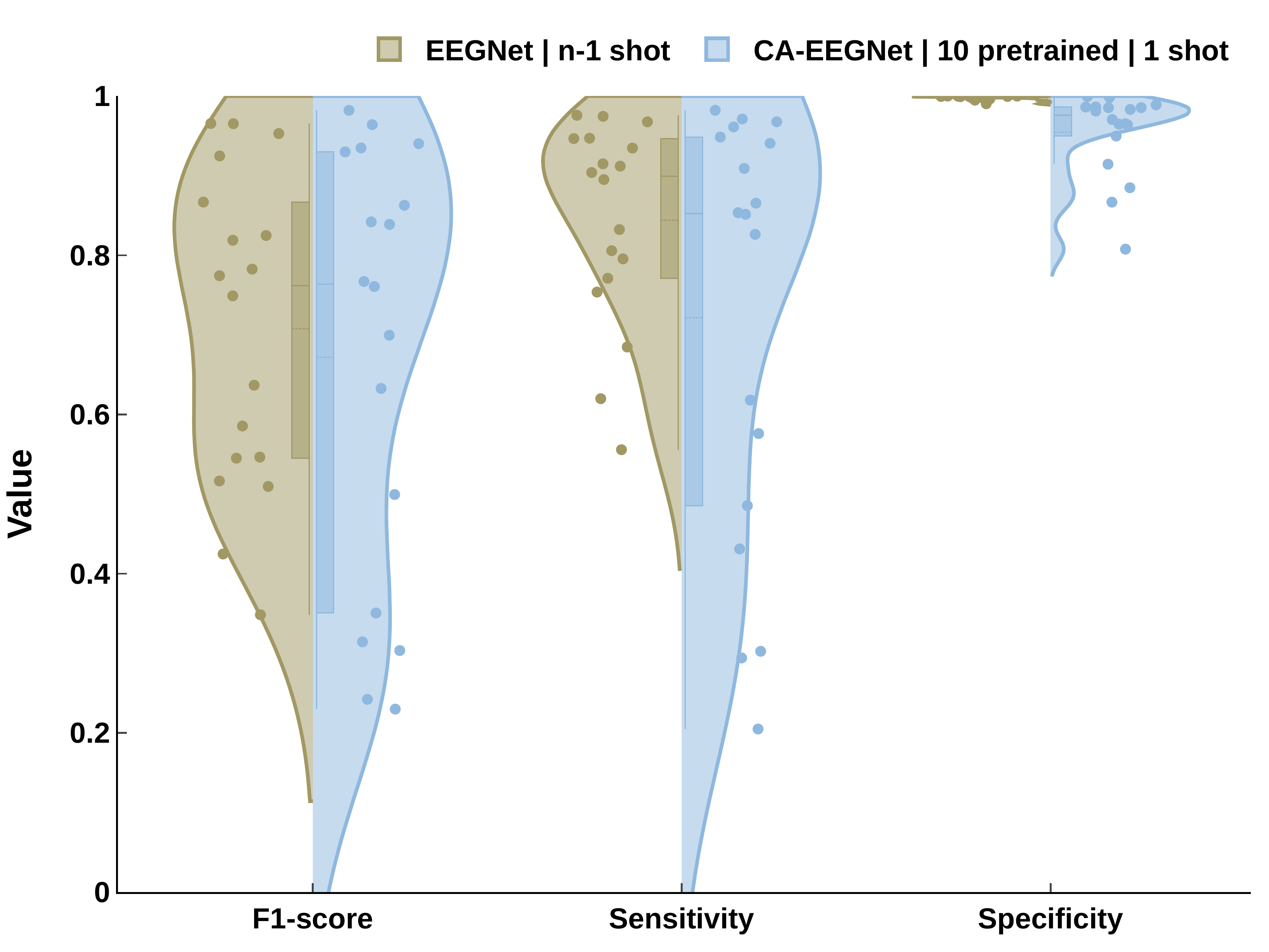}
    \caption{Subject-independent \namee\ (pre-trained on 15 subjects) surpasses subject-dependent EEGNet in F1-score, even when fine-tuned on a single seizure.}
    \label{fig:results_long_eegnet}
\end{subfigure}\hfill
\begin{subfigure}{.45\linewidth}
    \centering
    \includegraphics[width=\linewidth]{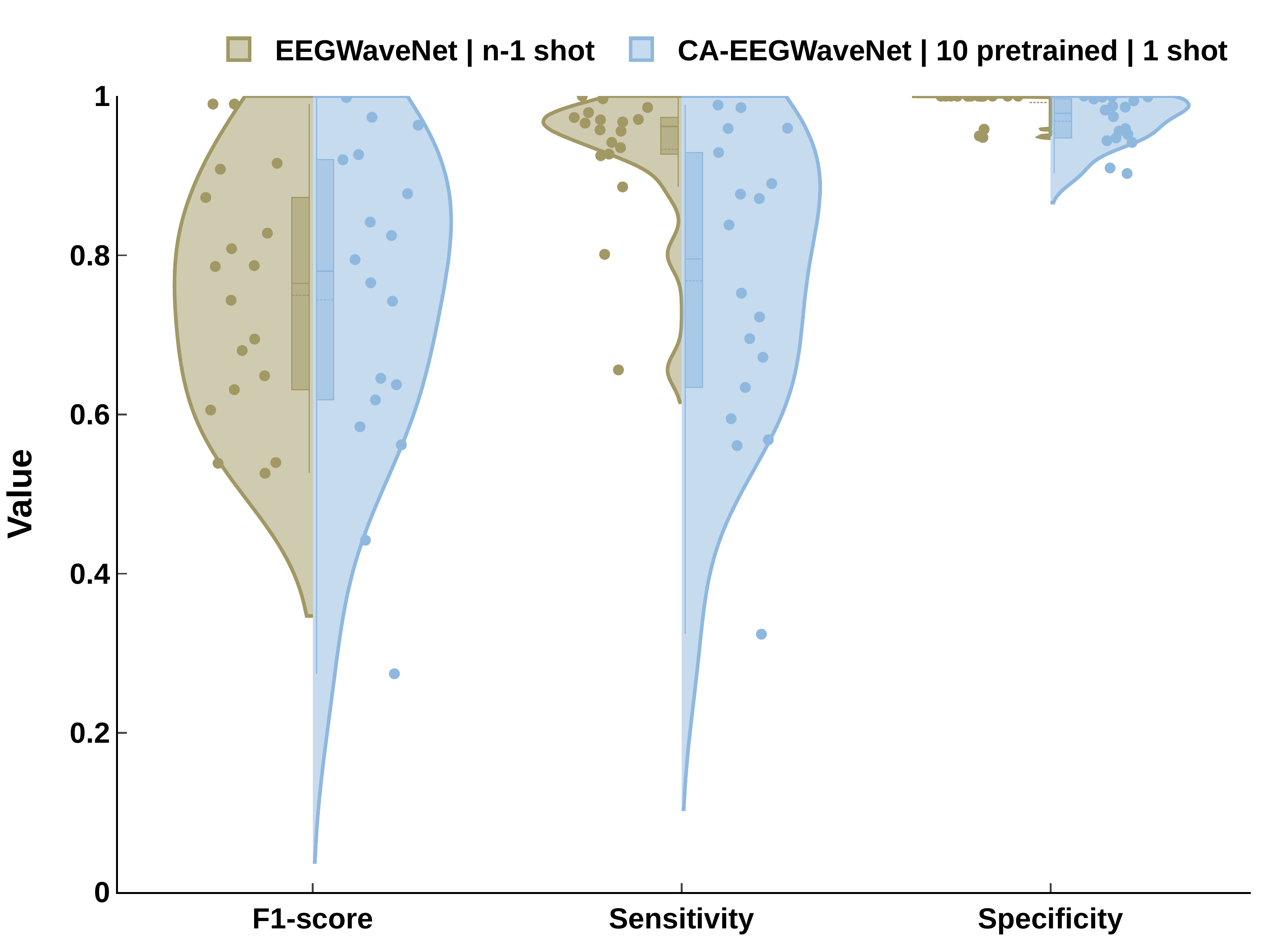}
    \caption{Subject-independent \namew\ (pre-trained on 10 subjects) surpasses subject-dependent EEGWaveNet in F1-score, even when fine-tuned on a single seizure.}
    \label{fig:results_long_eegwavenet}
\end{subfigure}
\caption{Seizure classification performance of our CA models on the Long-term SWEC iEEG Dataset. The CA models are fine-tuned on a single seizure of the subject (LABOC), compared with all but one seizures for the baseline models (LOOC). The mean (dashed line), median (solid line), and quartile boxes are visible.}
\end{figure*}

\begin{figure*}[ht]
\begin{subfigure}[t]{.45\linewidth}
    \centering
    \includegraphics[width=\linewidth]{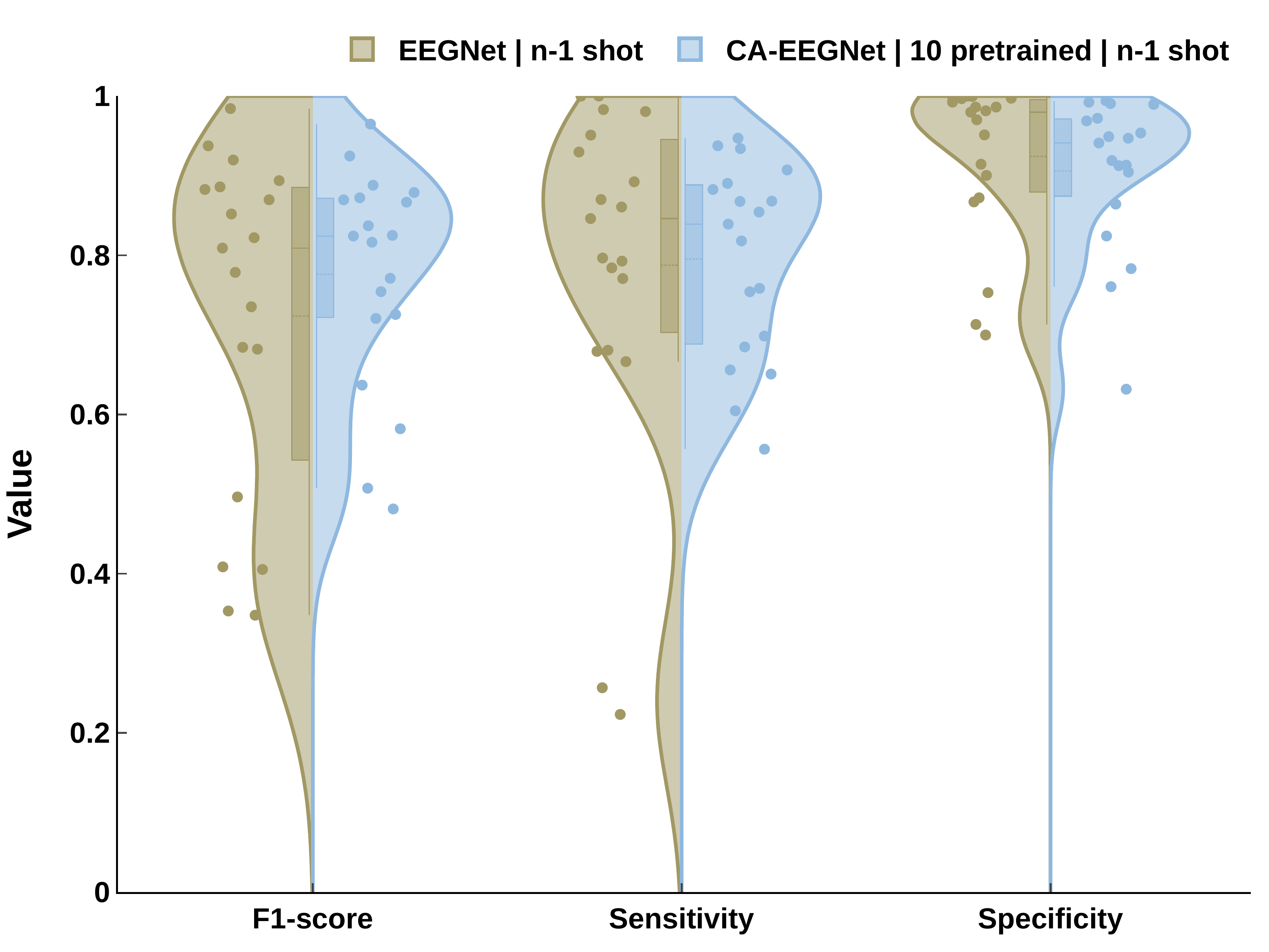}
    \caption{Subject-independent \namee\ (pre-trained on 10 subjects) surpasses subject-dependent EEGNet in F1-score.}
    \label{fig:results_short_eegnet}
\end{subfigure}\hfill
\begin{subfigure}[t]{.45\linewidth}
    \centering
    \includegraphics[width=\linewidth]{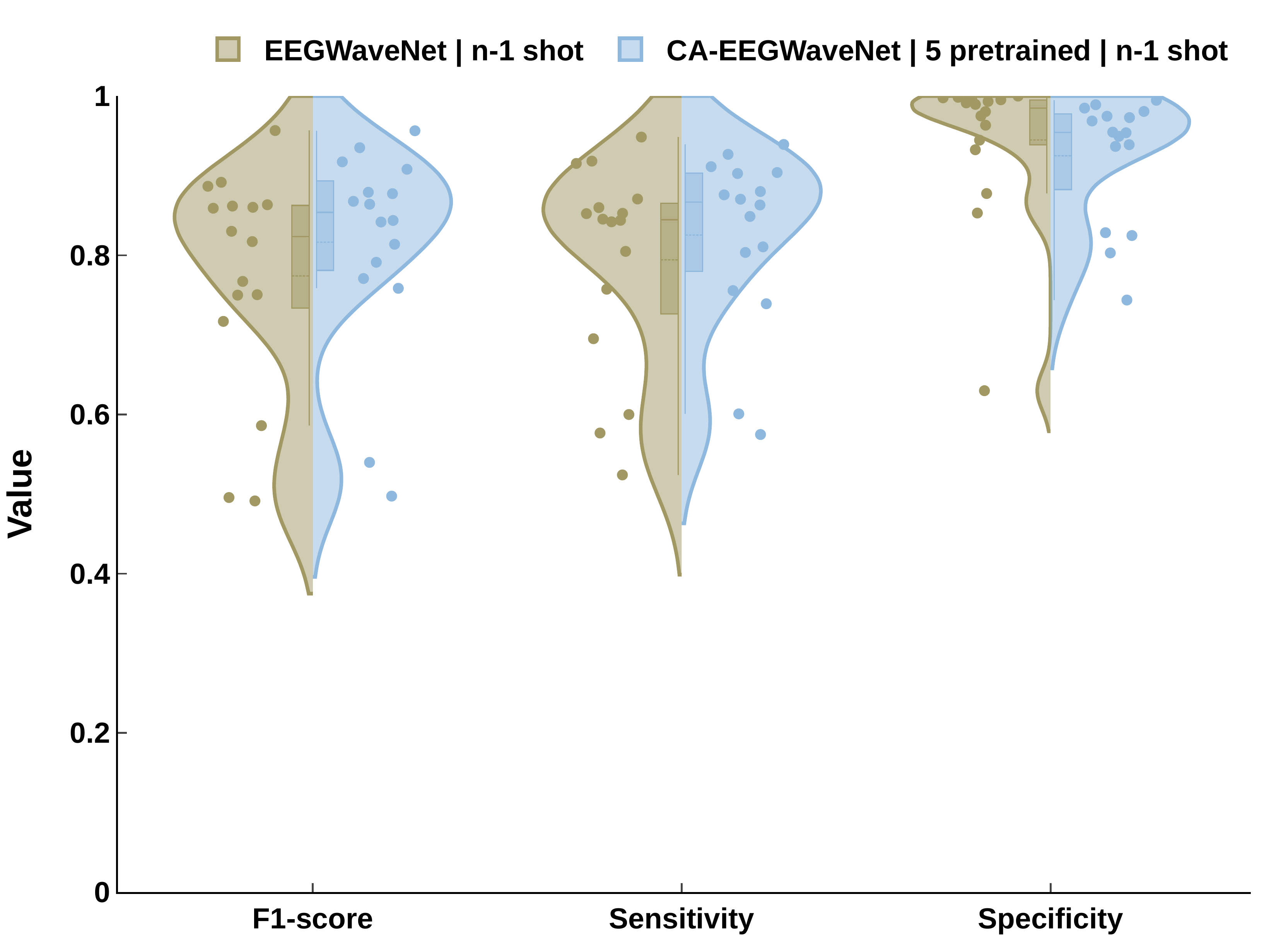}
    \caption{Subject-independent \namew\ (pre-trained on 5 subjects) surpasses subject-dependent EEGWaveNet in F1-score.}
    \label{fig:results_short_eegwavenet}
\end{subfigure}
\caption{Seizure classification performance of our CA models on the Short-term SWEC iEEG Dataset. The mean (dashed line), median (solid line), and quartile boxes are visible.}
\label{fig:results_short_all}
\end{figure*}

\subsection{Datasets}

\textbf{Short-term SWEC iEEG~\cite{Burrello2018}.} This short-term iEEG dataset contains 16 subjects, 14 hours of recording, and 104 ictal events. Each ictal event is accompanied by 3 minutes of pre-ictal and 3 minutes of post-ictal signal. The iEEG signals were recorded intracranially with a sampling rate of either 512\,Hz or 1024\,Hz. We downsample all signals to 512\,Hz for training all our models, both CA and baselines. The signals were median-referenced and band-pass filtered between 0.5 and 120\,Hz using a fourth-order Butterworth filter, both in a forward and backward pass. All the recordings were inspected by an expert neurologist for identification of seizure onsets and offsets, and to remove channels corrupted by artifacts.

\textbf{Long-term SWEC iEEG~\cite{Burrello2019}.} This long-term iEEG dataset contains 18 subjects, 2321 hours of recording, and 244 ictal events. Each subject is recorded continuously from implantation to explantation, regardless of the number of ictal events.  The recording, post-processing, and seizure analysis setup are identical to the Short-term SWEC iEEG dataset.

\subsection{Long-term dataset curation}

In order to effectively train many models on the Long-term SWEC iEEG dataset, we must select a subset of the non-ictal periods for each subject. We find that a selection strategy based on the Delta power of the signal is the most effective in terms of raw final performance. In particular, we compute the power spectrum decomposition of the entire signal for each subject in 4 second windows. We then aggregate the Delta power, thought to be most useful for seizure detection, into 5 bins. Finally, we take 20 minutes of signals from each bin randomly. These 100 minutes of signal, together with all the seizures, form the available pre-training dataset. 

\section{Results}

\subsection{Long-term dataset}

We use the Long-term SWEC iEEG dataset to evaluate our CA models against a set of baselines in a clinically-relevant scenario.
The recordings in this dataset are continuous and contain a large amount of variety in both ictal and interictal signals. Training follows the schema outlined in Section~\ref{sec:training}, specifically regarding the curation of the dataset for the pre-training phase. In the testing phase, each seizure is presented to the model as-is together with one hour of pre-ictal and one hour of post-ictal recordings.

Given the large amount of pre-training data at our disposal, we fine-tune our CA architectures using the LABOC scheme. This means our models receive notably less subject-specific data than the baselines, but still overall a larger amount of data from other subjects as well.

The results of \namee\ pre-trained on 15 subjects against the baseline models are shown in Fig.~\ref{fig:results_long_eegnet}. Remarkably, \namee\ achieves a median F1-score of $0.79$ against $0.74$ of the baseline, even when trained with only a fraction of the subject-specific data available to the baselines. This result confirms the advantage that increasing the pre-training dataset confers to the CA classifier, and highlights the generalization capabilities of our model.

We showcase the composability of our architecture and the advantages it confers by training \namew. EEGWaveNet has been shown to have increased performance on the task at hand, so we expect the same improvement to transfer to \namew. We then pre-train \namew\ on fewer subjects than \namee, namely 10 instead of 15, using the Long-term dataset and compare its performance with the baseline EEGWaveNet. The results in Fig.~\ref{fig:results_long_eegwavenet} indicate that \namew\ surpasses EEGWaveNet with a median F1-score of $0.78$ against $0.76$.

Therefore, we have shown that our architecture is indeed composable and that improvements in the Encoder models translate to improvements of the overarching model. Moreover, our architecture also scales in the size of the dataset.

Overall, the results \namee\ and \namew\ achieve on the Long-term SWEC iEEG dataset validate our two-phase deployment. First, the base, non-personalized model can be pre-trained on a large amount of non-subject-specific data which is already available, either publicly or on an institution-by-institution basis. This initial expense, both in time and computation power required, does not need to be repeated and is offset by future use. In particular, fine-tuning \namee\ and \namew\ requires 5$\times$ fewer epochs than fully training a baseline model. Second, a personalized model can be quickly fine-tuned using limited amounts of subject-specific recordings, which can be more rapidly obtained in a few hours. As the fine-tuning time is minimal and mostly constant, the personalized model can be quickly deployed within a few minutes to hours and be put to immediate use.

\subsection{Short-term dataset}

The Short-term SWEC iEEG dataset allows us to evaluate \namee\ and \namew\ against the baselines in a simpler scenario, where we also ablate the various components as shown in Section~\ref{sec:abl}. Each seizure in the Short-term dataset is accompanied by 3 minutes of pre-ictal and 3 minutes of post-ictal recording, making this a relatively balanced dataset and an ideal testing scenario. Every result is collected following the pre-training and fine-tuning schema presented in Section~\ref{sec:training}. 

Fig.~\ref{fig:results_short_eegnet} shows the performance of \namee\ pre-trained on 10 subjects and fine-tuned with LOOC. \namee\ surpasses its baseline EEGNet with a median F1-score of $0.82$ against $0.80$. In turn, specificity is lower, as the model is exposed to more data and this can lead to more false positives. Overall, \namee\ offers better performance and faster training with $\sim$300K less parameters.

We present the results of \namew\ on the seizure detection task in Fig.~\ref{fig:results_short_eegwavenet}. Already with 5 subjects pre-training \namew\ outperforms its baseline both in F1-score (median of $0.85$ against $0.82$) and sensitivity (median of $0.87$ against $0.84$). We also confirm once more that EEGWaveNet is a better Encoder model than EEGNet, and that its performance transfers to \namew\ as well. 

Overall, this indicates that our architecture can be a plug-and-play replacement for different models while bringing superior performance, faster training, and better generalisation capabilities.

\subsection{Component ablation}\label{sec:abl}

We evaluate the importance of each part of our pipeline separately, both in term of training routine and model components. Table~\ref{tab:ablation} presents a summary of our results.

\begin{table}[th]
\centering
\begin{tabular}{l|lll}
                                                   & F1-score    & Sensitivity & Specificity  \\ 
\hline
\rowcolor[rgb]{0.753,0.753,0.753} EEGNet           & 0.72 (0.21) & 0.79 (0.21) & \textbf{0.92} (0.10)  \\
CA-EEGNet                                          & \textbf{0.78} (0.13) & \textbf{0.79} (0.12) & 0.91 (0.10)  \\
\rowcolor[rgb]{0.753,0.753,0.753} w/o pre-training & 0.47 (0.27) & 0.51 (0.33) & 0.84 (0.22)  \\
w/o fusion                                         & 0.45 (0.29) & 0.50 (0.30) & 0.85 (0.13)  \\
\rowcolor[rgb]{0.753,0.753,0.753} w/o memory       & 0.51 (0.22) & 0.50 (0.22) & 0.91 (0.07) 
\end{tabular}
\caption{Summary of the ablation results for each component on the short-term dataset. The mean (std) F1-Score, sensitivity, and specificity are reported.}
\label{tab:ablation}
\end{table}

\textbf{Pre-training phase.} To evaluate the effectiveness of pre-training, we train \namee\ following the same pipeline as for EEGNet, i.e., without pre-training. As expected, pre-training is very beneficial to the model as it greatly improves the variance in the training samples and increases performance across the board.

\textbf{Fusion component.} To evaluate the impact of the Fusion component, we repeat the evaluation pipeline on a \namee\ without HRR Fusion. To achieve the combination of the channels we simply take the mean of all channels. The performance is drastically reduced (0.45 vs 0.78), indicating that our novel learnable HRR Fusion is necessary to fuse the channels.

\begin{figure*}[t]
    \centering
    \includegraphics[width=.8\textwidth]{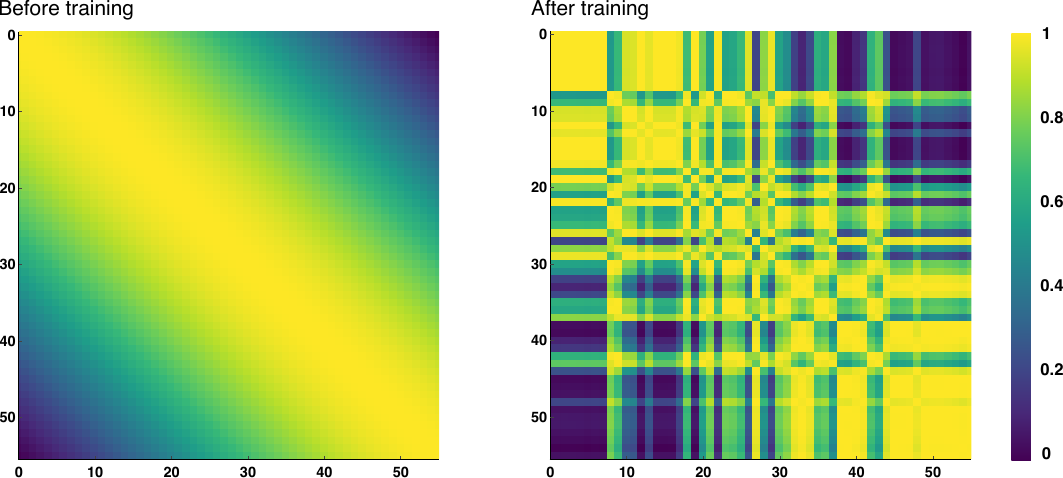}
    \caption{The Fusion component of \namee\ learns the spatial structure of the signal as training progresses, visualized through the cosine similarity between the generated key vectors. On the left, the component is initialized uniformly at the beginning of training and the relationship between the channels is simply linear. On the right, at the end of training, the relationship between the channels is more complex, representing a biologically plausible view of the connection between different electrodes and brain areas.}
    \label{fig:spatial_structure}
\end{figure*}

\textbf{Memory component.} To evaluate the impact of the Memory component, we repeat the evaluation pipeline on a \namee\ without TCN Memory. We simply feed the Fusion component to an MLP for final classification. The performance is drastically reduced (0.51 vs 0.78), indicating that the Memory component plays a significant role in the performance of the CA architecture.

The full details on the component ablation experiments can be found in Section~\ref{sec:app_abl}.

\subsection{Subject scaling}

Our CA architecture has the capability of ingesting any number of subjects for pre-training, while the baseline models need to be re-trained from scratch for every subject. Therefore, we can pre-train \namee\ and \namew\ with arbitrarily large datasets to increase their generalisation capabilities. At the same time, our CA models are limited in size, which might lead to sub-optimal fitting.

\begin{figure}[htb]
    \centering
    \includegraphics[width=.9\linewidth]{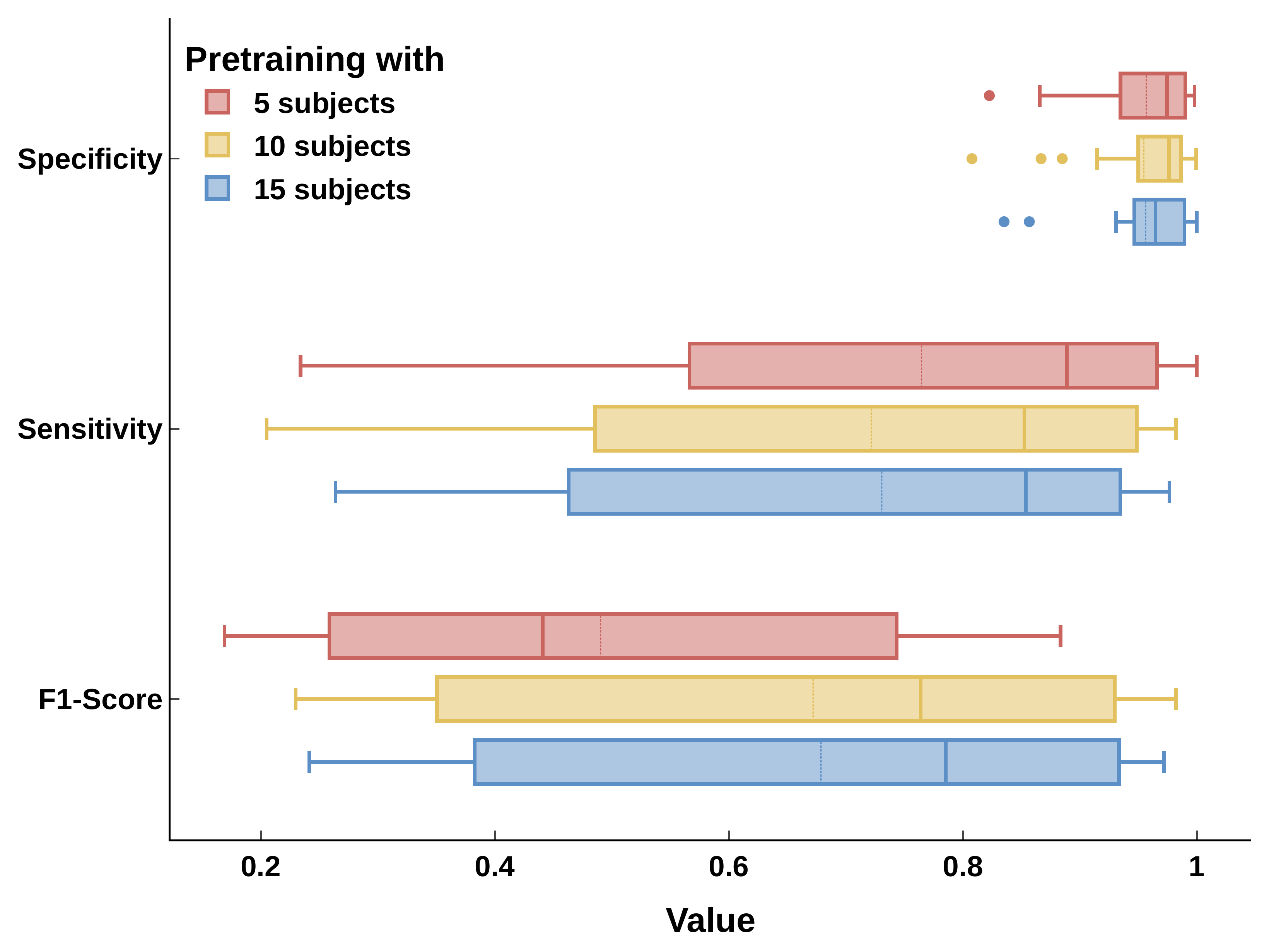}
    \caption{Performance of \namee\ when pre-trained with increasing number of subjects on the Long-term SWEC iEEG dataset. Performance increases up to 15 subjects but slows down, indicating that the scaling advantage is saturating. The mean (dashed line), median (solid line), and quartile boxes are visible.}
    \label{fig:results_scaling_subjects}
\end{figure}

We evaluate this behaviour by pre-training \namee\ on the Long-term dataset with 5, 10, and 15 subjects. The results in Fig.~\ref{fig:results_scaling_subjects} confirm our hypothesis. Performance increases by increasing the number of subjects from 5 to 15, yielding the state-of-the-art results shown previously in Fig.~\ref{fig:results_long_eegnet}. However, the scaling performance is not linear, indicating more powerful Encoder models are likely necessary to fully take advantage of the increased dataset size.

\subsection{Fusion component}

The Fusion component is the key to the channel-adaptive behavior of our CA architecture, therefore we separately evaluate how it learns the spatial structure of the signal. At the beginning of training, the HRR combination vector is uniformly initialized for each subject, to represent the simplest possible inductive bias on the spatial structure. This initialization implies that nearby channels are more strongly related. As training progresses, we expect the Fusion component to autonomously form a more nuanced understanding of the relationship between the channels. 

We encode this relationship as the cosine similarity between each channel's key vectors and observe how it evolves during training. Fig.~\ref{fig:spatial_structure} clearly shows that \namee\ does in fact learn information about the spatial structure of the signal, e.g., by forming blocks of similar channels that presumably belong to the same electrode. At the same time, the connection between some distant channels is also made stronger, which is compatible with a biologically plausible explanation of different areas of the brain being more strongly connected, regardless of the distance~\cite{Abela2014}.

\section{Conclusion}

In this work, we introduce a new composable channel-adaptive classifier, which is able to process iEEG signals from any subject regardless of their clinical setup, thanks to a novel learnable holographic channel fusion scheme. We show that our model can be deployed as a one-to-one replacement of current state-of-the-art deep learning models for iEEG processing. Compared to other solutions capable of channel-adaptive classification (e.g., Transformers~\cite{Yuan2024}), our models are 3 to 4 orders of magnitude smaller ($\sim$1.2B parameters vs $\sim$1M), making them more practical for clinical usage. Our composable architecture is not limited to the models we have showcased here, and can in principle be applied to any Encoder. This will allow future models to also become channel-adaptive and take advantage of larger training datasets. The resulting CA models are more performant and faster to train and deploy, and depending on the design of the Encoder architecture, they are also smaller in size and more lightweight.

\appendices

\renewcommand{\thefigure}{A\arabic{figure}} 

\section{Additional results}

\subsection{LABOC fine-tuning on the Short-term dataset}

\begin{figure}[h]
    \centering
    \includegraphics[width=.9\linewidth]{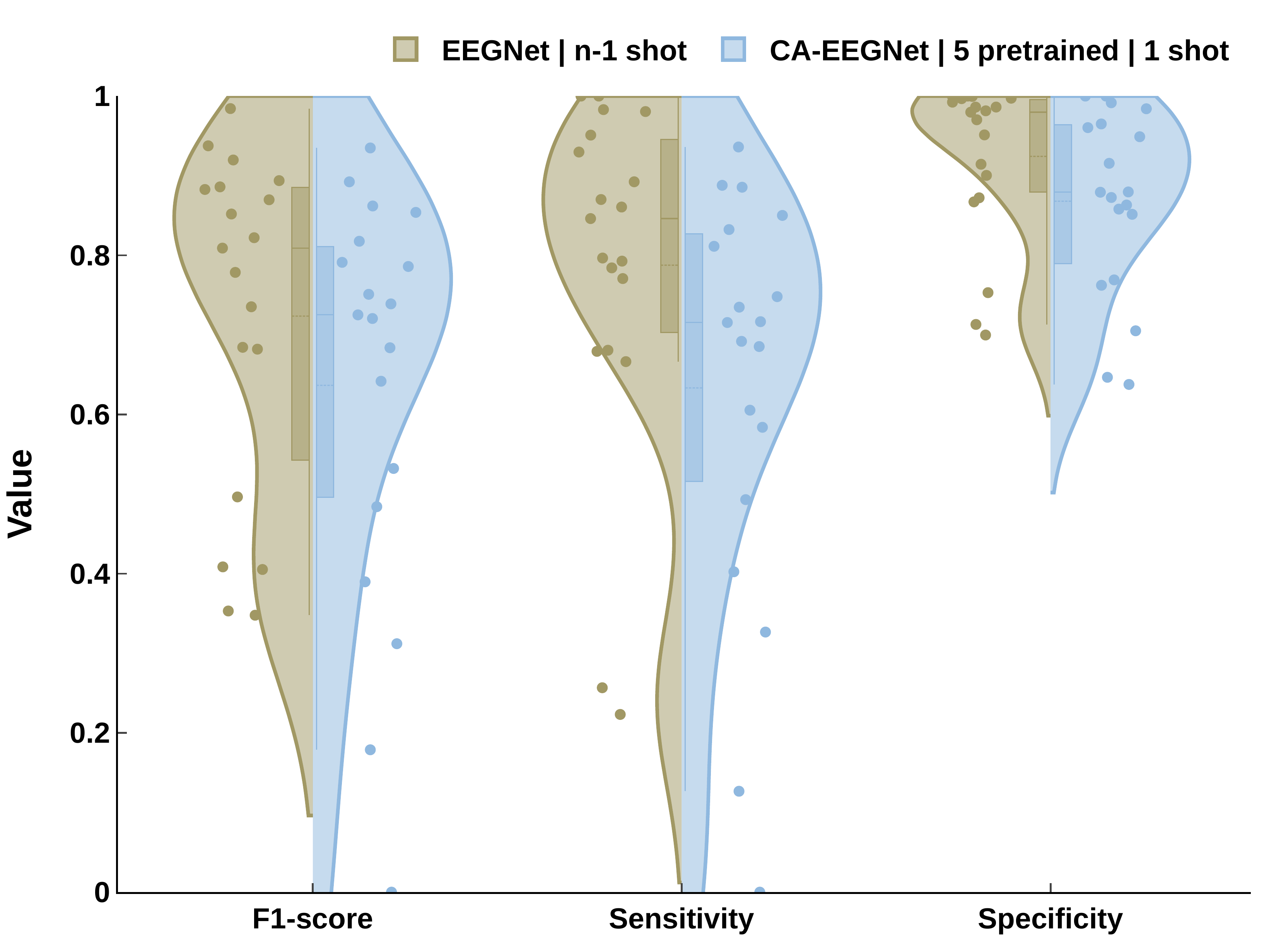}
    \caption{Performance of \namee\ when fine-tuned with a single seizure of the specific subject (LABOC) on the Short-term dataset. The mean (dashed line), median (solid line), and quartile boxes are visible.}
    \label{fig:laboc_short}
\end{figure}

We evaluate the effect of the fine-tuning stage on CA models with the Short-term SWEC iEEG dataset. Fig.~\ref{fig:laboc_short} indicates that a single seizure in the pre-training phase is not sufficient alone to generalize effectively to the others, due to the limited size of the dataset. Therefore, training with the same amount of data as the baseline model yields the best results.

\subsection{LOOC fine-tuning on the Long-term dataset}

\begin{figure}[h]
    \centering
    \includegraphics[width=.9\linewidth]{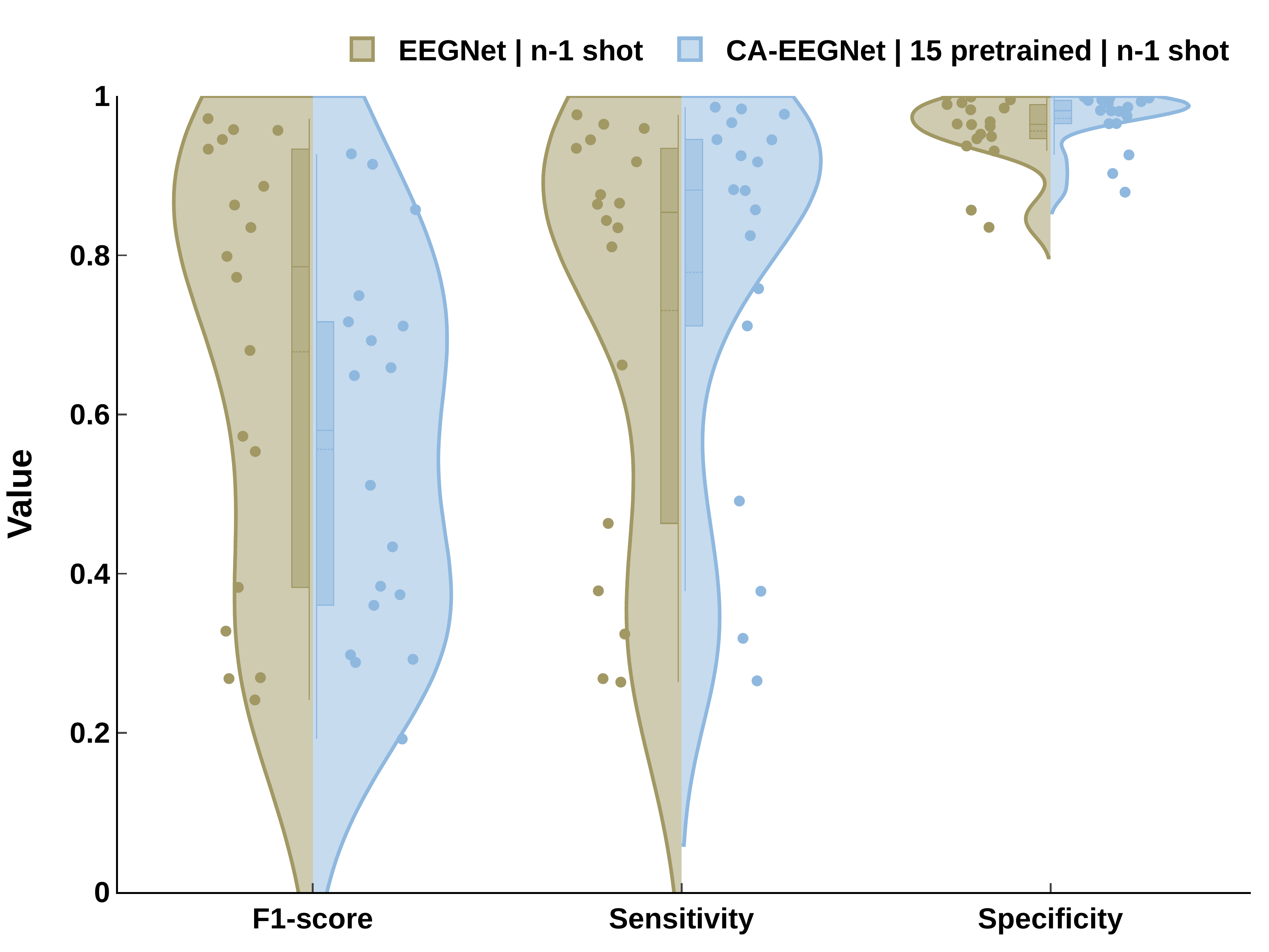}
    \caption{Performance of \namee\ when fine-tuned with a single seizure of the specific subject (LOOC) on the Long-term dataset. The mean (dashed line), median (solid line), and quartile boxes are visible.}
    \label{fig:looc_long}
\end{figure}

We evaluate the effect of the fine-tuning stage on CA models with the Long-term SWEC iEEG dataset. In contrast to the previous results, here training with too many seizures leads to sub-optimal fitting, as shown in Fig.~\ref{fig:looc_long}. For more details about this behavior see Section~\ref{sec:results_scaling_short}. Therefore, the best fine-tuning regime is with a single seizure to adapt the Fusion 
component while maintaining generality. 

\subsection{Subject scaling on the Short-term dataset}\label{sec:results_scaling_short}

We evaluate the subject scaling performance on the Short-term SWEC iEEG Dataset. In contrast to the Long-term dataset, we see a saturation as the number of training subjects increases. We hypothesize this is due to the lower variety in the samples of the Short-term dataset, which might hinder performance due to over-fitting and lower generality of the dataset.

\begin{figure}[h]
    \centering
    \includegraphics[width=.9\linewidth]{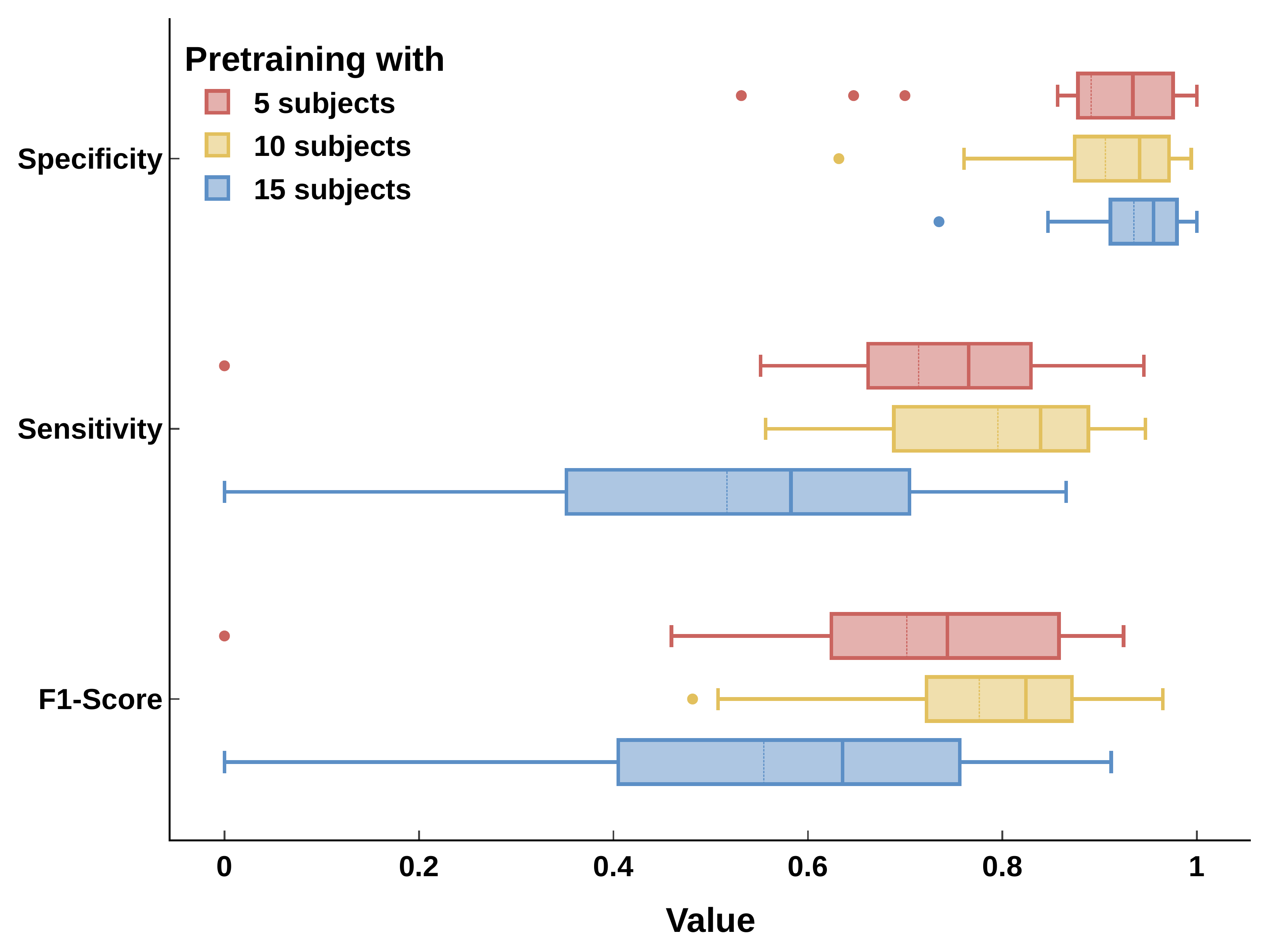}
    \caption{Performance of \namee\ when pre-trained with increasing number of subjects on the Short-term SWEC iEEG dataset. Performance increases up to 10 subjects and then regresses, indicating that network capacity has been reached. The mean (dashed line), median (solid line), and quartile boxes are visible.}
    \label{fig:results_scaling_subjects_short}
\end{figure}

We evaluate this behaviour by pre-training \namee\ on the Short-term dataset with 5, 10, and 15 subjects. The results in Fig.~\ref{fig:results_scaling_subjects_short} confirm our hypothesis. Performance initially increases by increasing the number of subjects from 5 to 10, yielding the state-of-the-art results shown previously in Fig.~\ref{fig:results_short_all}. However, at 15 pre-training subjects performance starts to regress. This suggests that the network capacity has been reached, and further training is no longer useful and even detrimental. By increasing the variance in the training samples, e.g., by using the Long-term dataset as shown in Fig.~\ref{fig:results_scaling_subjects}, we limit the over-fitting and can train with more data. Another approach to mitigate over-fitting would be to use larger Encoder and Memory components.

\subsection{Ablations}\label{sec:app_abl}

We present here the full details about the performance of CA-architectures with and without the different components.

\begin{figure}[h]
    \centering
    \includegraphics[ width=.9\linewidth]{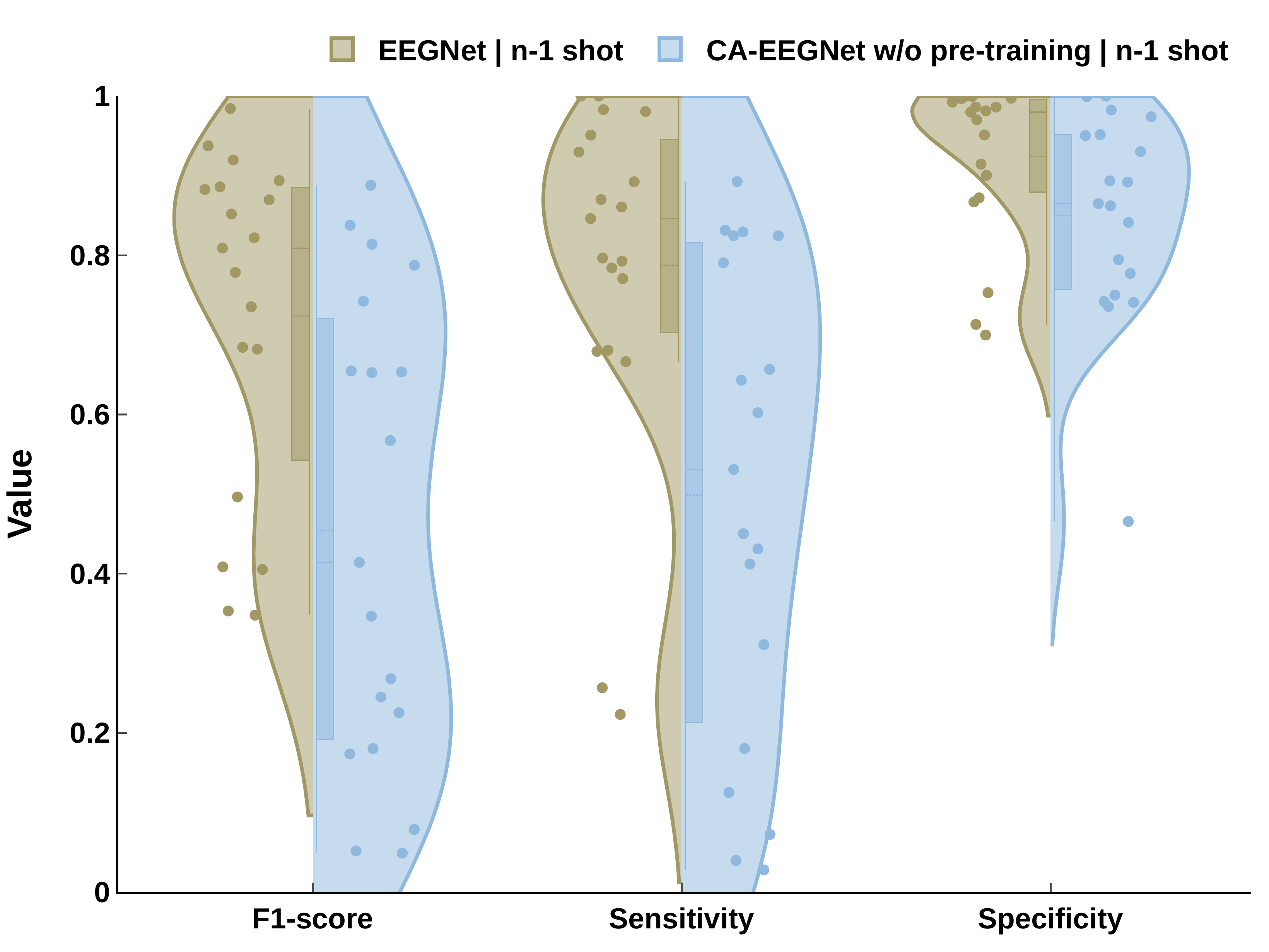}
    \caption{Seizure classification performance of \namee\ without pre-training.}
    \label{fig:nopretrain}
\end{figure}

\textbf{Pre-training phase.} The results are shown in Fig.~\ref{fig:nopretrain}. 

\begin{figure}[h]
    \centering
    \includegraphics[width=.9\linewidth]{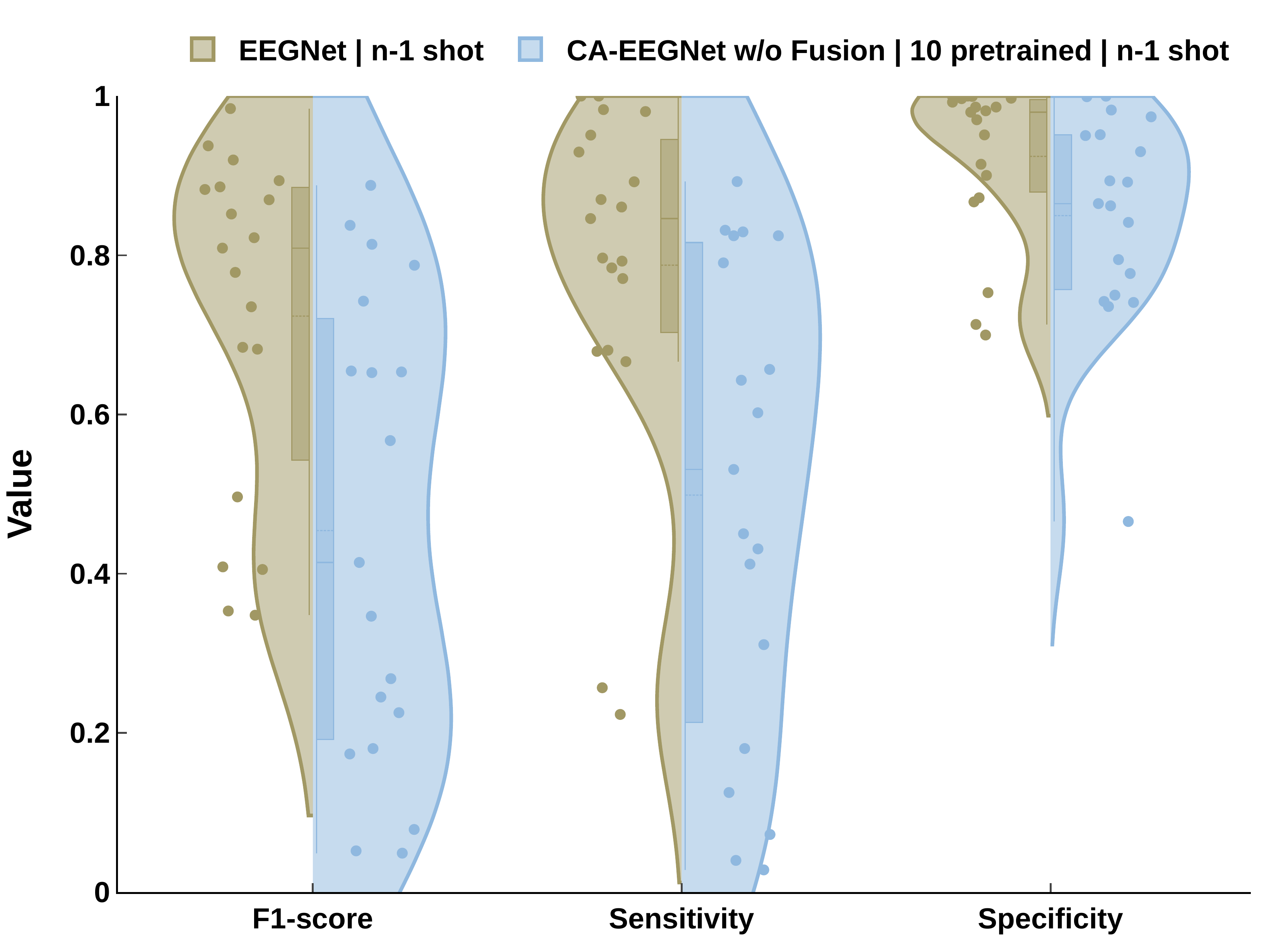}
    \caption{Seizure classification performance of \namee\ without HRR Fusion component.}
    \label{fig:nohrr}
\end{figure}

\textbf{Fusion component.} The results are shown in Fig.~\ref{fig:nohrr}.

\begin{figure}[h]
    \centering
    \includegraphics[width=.9\linewidth]{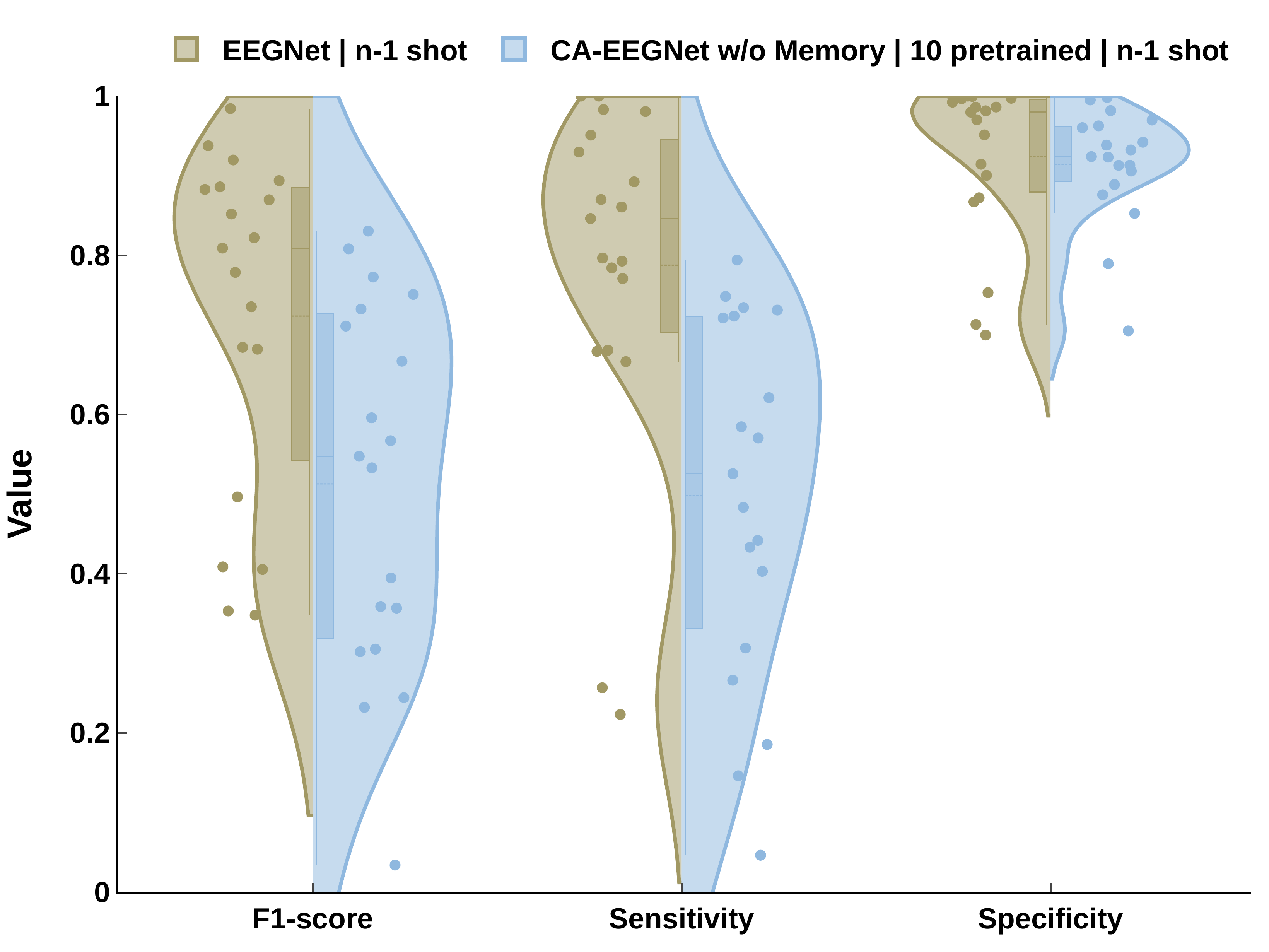}
    \caption{Seizure classification performance of \namee\ without TCN Memory component.}
    \label{fig:notcn}
\end{figure}

\textbf{Memory component.} The results are shown in Fig.~\ref{fig:notcn}.

\section{Fractional power encoding}\label{sup:fpe}

Vector symbolic architectures (VSAs)~\cite{Kanerva2009} are symbolic models that use vector representations to build semantically meaningful subspaces of $\mathbb{R}^d$. The vectors used in VSAs are usually drawn randomly and i.i.d., such that a large number of them can be drawn without interfering with each other. This translates to the vectors in the collection having low similarity (e.g., cosine similarity) and is guaranteed by the concentration of measure as $d$ becomes larger. To build the aforementioned subspaces, the vectors are combined through two elementary operations: binding and bundling. On one hand, binding serves to build an output vector which is dissimilar to the input vectors used in its creation. On the other hand, bundling combines multiple vectors into one which is similar to all its components.

VSA models are characterised mainly by these two properties: the random vectors and the elementary operations. In this work, we use fractional power encoding (FPE)~\cite{Plate1992} together with holographic reduced representations (HRR)~\cite{Plate1995}. We now analyse the suitability of this VSA model for the Fusion component of our channel-adaptive classifier.

We use FPE to efficiently track the relationship between the spatial channels of an iEEG signal and generate appropriate symbolic vectors. Consistent with the HRR paradigm, we also choose circular convolution ($\circledast$) to be our binding operation, and simple summation to be our bundling operation. Circular convolution can be efficiently implemented by considering it as multiplication in Fourier space, such that
\begin{equation}
    \bm{v} \circledast \bm{w} = \mathcal{F}^{-1}(\mathcal{F}(\bm{v}) \odot \mathcal{F}(\bm{w})),
\end{equation}
where $\odot$ is elementwise multiplication. Using FFT, $\circledast$ becomes computationally very lightweight.

Next, we can apply this binding operation to generate FPE key vectors. Given a random vector $\bm{v}$ and an angle $r$, we can rotate $\bm{v}$ by $2\pi \times r$ radians by self-binding $r$ times as follows
\begin{equation}
    \text{rot}(\bm{v}, r) = \bm{v} \underbrace{\circledast \dots \circledast}_{\text{$r$ times}} \bm{v} = \mathcal{F}^{-1}(\mathcal{F}(\bm{v})^{r}).
\end{equation}
We want to ensure that each rotated vector is only different in its angle to another, to avoid unduly biasing the overall model towards one specific angle. However, rot() does not necessarily preserve the norm of $\bm{v}$. To fix this issue, we must first make $\bm{v}$ unitary, i.e. normalize it such that
\begin{equation}
    \norm{\mathcal{F}(\bm{v})_i} = 1 \quad \forall i.
\end{equation}

We now have the required structure to fuse the channels and preserve their spatial relationship. As before, $\bm{m}_{\text{ch}}$ is a vector containing angles, each associated to a specific channel. To avoid possible issues with different basis vectors $\bm{v}$, as $ \text{rot}(\cdot, 0)$ produces the same result for any vector, we constrain $\bm{m}_{\text{ch}}$ to have values from $1$ to $2$. We then perform subtraction by $1$ in rot to recover the angle interpretation. The outputs of the encoder component are the channel value vectors $\bm{p}_{i}$, which are first bound with the FPE channel key vectors and finally bundled as follows
\begin{equation}
    \bm{f} = \sum_{i} \bm{p}_{i} \circledast \text{rot}(\bm{v}, \bm{m}_{\text{ch}}^i).
\end{equation}
We get one such fused vector for each time window in the signal. The convolutional binding operator is distributive with respect to addition, meaning that similar key vectors constructively interfere. In contrast, dissimilar key vectors produce dissimilar results, such that the dissimilar channels do not interact in the bundled vector. Moreover, each subject has its own $\bm{m}_{\text{ch}}$, since the spatial configuration of the electrodes is unique.

The resulting kernel from this VSA realisation approximates the sinc function, therefore the relationship between the channels decays symmetrically as the difference in the angles gets larger. In addition, $\bm{m}_{\text{ch}}$ is learned by the network and translates well to the spatial structure of the electrodes. This can be seen by plotting the cosine similarity of the FPE key vectors at the beginning and at the end of training (see Fig.~\ref{fig:results_scaling_subjects} in the main Manuscript). In summary, this schema provides a maximally economical (only one value per channel) representation of the spatial information content of an iEEG signal.

\section{Hardware details}

The hardware setup used for training the CA models includes 8$\times$ Nvidia A100 SXM4 80GB and an AMD EPYC 7763, with 1\;TiB of RAM (effective usage is lower). The software stack includes CUDA 11.8, cuDNN 8.7, PyTorch 2.1.1, and PyTorch Lightning 2.1.0. All models were implemented using PyTorch and trained using PyTorch Lightning.

\section*{Acknowledgment}

This work is supported by the Swiss National Science foundation (SNF), grant no. 200800.

\ifCLASSOPTIONcaptionsoff
  \newpage
\fi

\bibliographystyle{IEEEtran}
\bibliography{IEEEabrv,parall}

\begin{thebibliography}{10}
\providecommand{\url}[1]{#1}
\csname url@samestyle\endcsname
\providecommand{\newblock}{\relax}
\providecommand{\bibinfo}[2]{#2}
\providecommand{\BIBentrySTDinterwordspacing}{\spaceskip=0pt\relax}
\providecommand{\BIBentryALTinterwordstretchfactor}{4}
\providecommand{\BIBentryALTinterwordspacing}{\spaceskip=\fontdimen2\font plus
\BIBentryALTinterwordstretchfactor\fontdimen3\font minus
  \fontdimen4\font\relax}
\providecommand{\BIBforeignlanguage}[2]{{%
\expandafter\ifx\csname l@#1\endcsname\relax
\typeout{** WARNING: IEEEtran.bst: No hyphenation pattern has been}%
\typeout{** loaded for the language `#1'. Using the pattern for}%
\typeout{** the default language instead.}%
\else
\language=\csname l@#1\endcsname
\fi
#2}}
\providecommand{\BIBdecl}{\relax}
\BIBdecl

\bibitem{Liu2017}
F.~Liu, S.~Wang, J.~Rosenberger, J.~Su, and H.~Liu, ``A sparse dictionary
  learning framework to discover discriminative source activations in {EEG}
  brain mapping,'' \emph{Proceedings of the AAAI Conference on Artificial
  Intelligence}, vol.~31, no.~1, 2017.

\bibitem{Song2020}
T.~Song, S.~Liu, W.~Zheng, Y.~Zong, and Z.~Cui, ``Instance-adaptive graph for
  eeg emotion recognition,'' \emph{Proceedings of the AAAI Conference on
  Artificial Intelligence}, vol.~34, no.~03, 2020.

\bibitem{Zhao2021}
L.-M. Zhao, X.~Yan, and B.-L. Lu, ``Plug-and-play domain adaptation for
  cross-subject eeg-based emotion recognition,'' \emph{Proceedings of the AAAI
  Conference on Artificial Intelligence}, vol.~35, no.~1, 2021.

\bibitem{Rajpurkar2022}
P.~Rajpurkar, E.~Chen, O.~Banerjee, and E.~J. Topol, ``{AI} in health and
  medicine,'' \emph{Nature Medicine}, vol.~28, no.~1, 2022.

\bibitem{Wei2022}
X.~Wei, A.~A. Faisal, M.~Grosse-Wentrup, A.~Gramfort, S.~Chevallier,
  V.~Jayaram, C.~Jeunet, S.~Bakas, S.~Ludwig, K.~Barmpas \emph{et~al.}, ``2021
  {BEETL} competition: Advancing transfer learning for subject independence and
  heterogenous {EEG} data sets,'' in \emph{NeurIPS 2021 Competitions and
  Demonstrations Track}, 2022.

\bibitem{Heinrichs2023}
F.~Heinrichs, M.~Heim, and C.~Weber, ``Functional neural networks: Shift
  invariant models for functional data with applications to {EEG}
  classification,'' in \emph{Proceedings of the 40th International Conference
  on Machine Learning (ICML)}, 2023.

\bibitem{Ho2023}
T.~K.~K. Ho and N.~Armanfard, ``Self-supervised learning for anomalous channel
  detection in eeg graphs: Application to seizure analysis,'' \emph{Proceedings
  of the AAAI Conference on Artificial Intelligence}, vol.~37, no.~7, 2023.

\bibitem{Casson2019}
A.~J. Casson, ``Wearable {EEG} and beyond,'' \emph{Biomedical Engineering
  Letters}, vol.~9, no.~1, 2019.

\bibitem{Arico2020}
P.~Aricò, N.~Sciaraffa, and F.~Babiloni, ``Brain–computer interfaces: Toward
  a daily life employment,'' \emph{Brain Sciences}, vol.~10, no.~3, 2020.

\bibitem{Kuhlmann2018}
L.~Kuhlmann, K.~Lehnertz, M.~P. Richardson, B.~Schelter, and H.~P. Zaveri,
  ``Seizure prediction — ready for a new era,'' \emph{Nature Reviews
  Neurology}, vol.~14, no.~10, 2018.

\bibitem{Craik2019}
A.~Craik, Y.~He, and J.~L. Contreras-Vidal, ``Deep learning for
  electroencephalogram ({EEG}) classification tasks: a review,'' \emph{Journal
  of Neural Engineering}, vol.~16, no.~3, 2019.

\bibitem{Yuan2024}
Z.~Yuan, F.~Shen, M.~Li, Y.~Yu, C.~Tan, and Y.~Yang, ``Brant-2: Foundation
  model for brain signals,'' \emph{arXiv preprint arXiv:2402.10251}, Feb. 2024.

\bibitem{Burrello2019}
A.~Burrello, L.~Cavigelli, K.~Schindler, L.~Benini, and A.~Rahimi, ``Laelaps:
  An energy-efficient seizure detection algorithm from long-term human {iEEG}
  recordings without false alarms,'' in \emph{Design, Automation {\&} Test in
  Europe Conference {\&} Exhibition ({DATE})}, 2019.

\bibitem{Wang2023}
C.~Wang, V.~Subramaniam, A.~U. Yaari, G.~Kreiman, B.~Katz, I.~Cases, and
  A.~Barbu, ``Brain{BERT}: Self-supervised representation learning for
  intracranial recordings,'' in \emph{The Eleventh International Conference on
  Learning Representations (ICLR)}, 2023.

\bibitem{Burrello2018}
A.~Burrello, K.~Schindler, L.~Benini, and A.~Rahimi, ``One-shot learning for
  {iEEG} seizure detection using end-to-end binary operations: Local binary
  patterns with hyperdimensional computing,'' in \emph{2018 IEEE Biomedical
  Circuits and Systems Conference (BioCAS)}, 2018.

\bibitem{Shoeb2010}
A.~Shoeb, ``{CHB-MIT} scalp {EEG} database,'' 2010.

\bibitem{Yilmaz2024}
G.~Yilmaz, A.~Seiler, O.~Chételat, and K.~A. Schindler, ``{Ultra-Long-Term-EEG
  Monitoring (ULTEEM)} systems: Towards user-friendly out-of-hospital
  recordings of electrical brain signals in epilepsy,'' \emph{Sensors},
  vol.~24, no.~6, 2024.

\bibitem{Plate1995}
T.~Plate, ``Holographic reduced representations,'' \emph{IEEE Transactions on
  Neural Networks}, vol.~6, no.~3, 1995.

\bibitem{Schirrmeister2017}
R.~T. Schirrmeister, J.~T. Springenberg, L.~D.~J. Fiederer, M.~Glasstetter,
  K.~Eggensperger, M.~Tangermann, F.~Hutter, W.~Burgard, and T.~Ball, ``Deep
  learning with convolutional neural networks for eeg decoding and
  visualization,'' \emph{Human Brain Mapping}, vol.~38, no.~11, 2017.

\bibitem{Lawhern2018}
V.~J. Lawhern, A.~J. Solon, N.~R. Waytowich, S.~M. Gordon, C.~P. Hung, and
  B.~J. Lance, ``{EEGNet}: a compact convolutional neural network for
  {EEG}-based brain{\textendash}computer interfaces,'' \emph{Journal of Neural
  Engineering}, vol.~15, no.~5, 2018.

\bibitem{Vilamala2017}
A.~Vilamala, K.~H. Madsen, and L.~K. Hansen, ``Deep convolutional neural
  networks for interpretable analysis of {EEG} sleep stage scoring,'' in
  \emph{2017 IEEE 27th International Workshop on Machine Learning for Signal
  Processing (MLSP)}, 2017.

\bibitem{Palotti2019}
J.~Palotti, R.~Mall, M.~Aupetit, M.~Rueschman, M.~Singh, A.~Sathyanarayana,
  S.~Taheri, and L.~Fernandez-Luque, ``Benchmark on a large cohort for
  sleep-wake classification with machine learning techniques,'' \emph{npj
  Digital Medicine}, vol.~2, no.~1, 2019.

\bibitem{Perslev2021}
M.~Perslev, S.~Darkner, L.~Kempfner, M.~Nikolic, P.~J. Jennum, and C.~Igel,
  ``{U-Sleep}: resilient high-frequency sleep staging,'' \emph{npj Digital
  Medicine}, vol.~4, no.~1, 2021.

\bibitem{Cho2020}
K.-O. Cho and H.-J. Jang, ``Comparison of different input modalities and
  network structures for deep learning-based seizure detection,''
  \emph{Scientific Reports}, vol.~10, no.~1, 2020.

\bibitem{Thuwajit2022}
P.~Thuwajit, P.~Rangpong, P.~Sawangjai, P.~Autthasan, R.~Chaisaen,
  N.~Banluesombatkul, P.~Boonchit, N.~Tatsaringkansakul, T.~Sudhawiyangkul, and
  T.~Wilaiprasitporn, ``{EEGWaveNet}: Multiscale {CNN}-based spatiotemporal
  feature extraction for {EEG} seizure detection,'' \emph{{IEEE} Transactions
  on Industrial Informatics}, vol.~18, no.~8, 2022.

\bibitem{Du2022}
Y.~Du, Y.~Xu, X.~Wang, L.~Liu, and P.~Ma, ``{EEG} temporal–spatial
  transformer for person identification,'' \emph{Scientific Reports}, vol.~12,
  no.~1, 2022.

\bibitem{Lee2022}
M.~Lee, L.~R.~D. Sanz, A.~Barra, A.~Wolff, J.~O. Nieminen, M.~Boly,
  M.~Rosanova, S.~Casarotto, O.~Bodart, J.~Annen, A.~Thibaut, R.~Panda,
  V.~Bonhomme, M.~Massimini, G.~Tononi, S.~Laureys, O.~Gosseries, and S.-W.
  Lee, ``Quantifying arousal and awareness in altered states of consciousness
  using interpretable deep learning,'' \emph{Nature Communications}, vol.~13,
  no.~1, 2022.

\bibitem{Moin2021}
A.~Moin, A.~Zhou, A.~Rahimi, A.~Menon, S.~Benatti, G.~Alexandrov, S.~Tamakloe,
  J.~Ting, N.~Yamamoto, Y.~Khan \emph{et~al.}, ``A wearable biosensing system
  with in-sensor adaptive machine learning for hand gesture recognition,''
  \emph{Nature Electronics}, vol.~4, no.~1, 2021.

\bibitem{Gayler2003}
R.~W. Gayler, ``Vector symbolic architectures answer jackendoff's challenges
  for cognitive neuroscience,'' in \emph{Proceedings of the ICCS/ASCS Joint
  International Conference on Cognitive Science (ICCS/ASCS 2003)}, 2003.

\bibitem{Kanerva2009}
P.~Kanerva, ``Hyperdimensional computing: An introduction to computing in
  distributed representation with high-dimensional random vectors,''
  \emph{Cognitive Computation}, vol.~1, no.~2, 2009.

\bibitem{Lea2016}
C.~Lea, R.~Vidal, A.~Reiter, and G.~D. Hager, ``Temporal convolutional
  networks: A unified approach to action segmentation,'' in \emph{Computer
  Vision--ECCV 2016 Workshops}, 2016.

\bibitem{Abela2014}
E.~Abela, C.~Rummel, M.~Hauf, C.~Weisstanner, K.~Schindler, and R.~Wiest,
  ``Neuroimaging of epilepsy: Lesions, networks, oscillations,'' \emph{Clinical
  Neuroradiology}, vol.~24, no.~1, 2014.

\bibitem{Plate1992}
T.~A. Plate, ``Holographic recurrent networks,'' in \emph{Advances in Neural
  Information Processing Systems}, S.~Hanson, J.~Cowan, and C.~Giles, Eds.,
  vol.~5, 1992.

\end{thebibliography}

\end{document}